\documentclass[letterpaper]{article} 
\usepackage{aaai24}  
\usepackage{times}  
\usepackage{helvet}  
\usepackage{courier}  
\usepackage[hyphens]{url}  
\usepackage{graphicx} 
\usepackage{natbib}  
\usepackage{caption} 
\usepackage{algorithm}
\usepackage{algorithmic}

\usepackage{newfloat}
\usepackage{listings}
\DeclareCaptionStyle{ruled}{labelfont=normalfont,labelsep=colon,strut=off} 
\floatstyle{ruled}
\newfloat{listing}{tb}{lst}{}
\floatname{listing}{Listing}
\usepackage{amsmath}
\usepackage{makecell}
\usepackage{multirow}
\usepackage{amssymb}

\title{Divide and Conquer: Hybrid Pre-training for Person Search}
\author{
    Yanling Tian\textsuperscript{\rm 1}, Di Chen\textsuperscript{\rm 1}, Yunan Liu\textsuperscript{\rm 1,2}, Jian Yang\textsuperscript{\rm 1}, Shanshan Zhang\textsuperscript{\rm 1}\thanks{Corresponding author.}
    \\
}
\affiliations{

    \textsuperscript{\rm 1}PCA Lab, Key Lab of Intelligent Perception and Systems for High-Dimensional Information of Ministry of Education,\\
    and Jiangsu Key Lab of Image and Video Understanding for Social Security, \\
    School of Computer Science and Engineering, Nanjing University of Science and Technology, Nanjing, China \\
    \textsuperscript{\rm 2} School of Artificial Intelligence, Dalian Maritime University \\
    
    \{yl.tian, dichen, liuyunan, shanshan.zhang, csjyang\}@njust.edu.cn
}

\usepackage{bibentry}

\begin{document}

\maketitle

\begin{abstract}
Large-scale pre-training has proven to be an effective method for improving performance across different tasks. Current person search methods use ImageNet pre-trained models for feature extraction, yet it is not an optimal solution due to the gap between the pre-training task and person search task (as a downstream task). Therefore, in this paper, we focus on pre-training for person search, which involves \emph{detecting} and \emph{re-identifying} individuals simultaneously. 
   Although labeled data for person search is scarce, datasets for two sub-tasks person detection and re-identification are relatively abundant. 
   To this end, we propose a hybrid pre-training framework specifically designed for person search using sub-task data only. It consists of a hybrid learning paradigm that handles data with different kinds of supervisions, and an intra-task alignment module that alleviates domain discrepancy under limited resources.
   To the best of our knowledge, this is the first work that investigates how to support full-task pre-training using sub-task data. 
   Extensive experiments demonstrate that our pre-trained model can achieve significant improvements across diverse protocols, such as person search method, fine-tuning data, pre-training data and model backbone. For example, our model improves ResNet50 based NAE by 10.3$\%$ relative improvement w.r.t. mAP. 
   Our code and pre-trained models are released for plug-and-play usage to the person search community (\url{https://github.com/personsearch/PretrainPS}).
\end{abstract}

\section{Introduction}
Person search aims to localize a person and identify the person from a gallery set of real-world uncropped scene images, which can be seen as a combined pedestrian detection and re-identification (re-ID) task. Person search is an extremely difficult problem because it requires optimising these two different sub-tasks in a unified framework, and even the optimisation objectives of the two sub-tasks are inconsistent. 

\begin{figure}[!t]
  \centering
  \includegraphics[width=\linewidth]{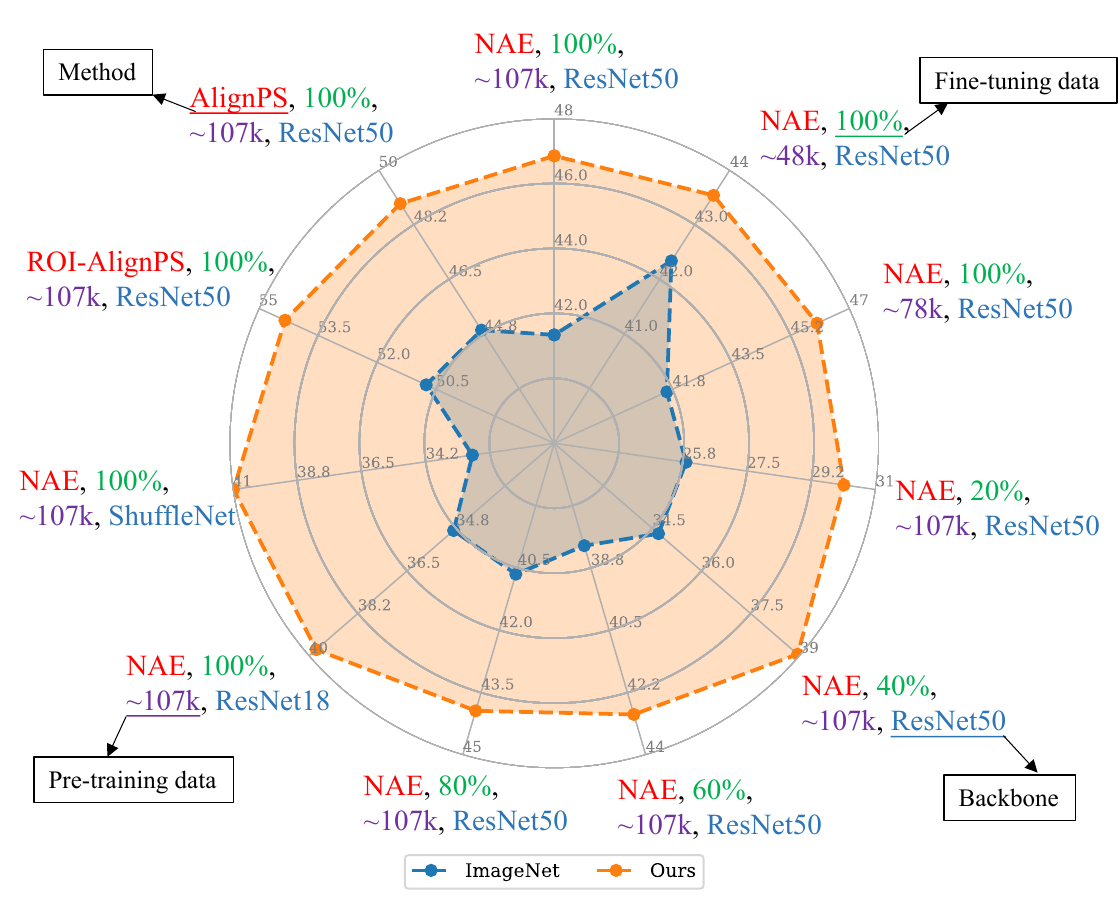}
  \caption{Performance gain using our pre-trained model instead of ImageNet model across different \textit{\textbf{methods, fine-tuning data, pre-training data, backbones}} on PRW dataset.}
  \label{fig:methods}
\end{figure}

 Most existing person search methods~\cite{CUHK,DAPS,CGPS,R-SiamNets,IJCV_NAE,PGA,SeqNet,TriNet,Alignps,Alignps_roi,tian2022} use ImageNet pre-trained models~\cite{ImageNet}, such as ResNet50~\cite{resnet}, as the initialization model for feature extraction. However, ImageNet pre-training, which learns classification-related knowledge, is limited in its applicability to downstream tasks, particularly when the target task is significantly different~\cite{rethink_imgnet}. The person search task, which includes a pedestrian detection task and a fine-grained identity classification task, requires prior knowledge in regression and there is a substantial domain gap between image classification and person re-ID tasks~\cite{luperson}. To address this issue, task-specific pre-training models are necessary. 

  Some recent works aim to design specific pre-training methods for different computer vision tasks~\cite{luperson,luperson_noisy,up_reid,detreg}.
 For example, DETR~\cite{detr} is pre-trained using a multi-task learning approach on the COCO training dataset (118,000 images) with object annotations for both object detection and object re-localization tasks;~\cite{action} propose a new model pre-trained on the proposed Kinetics dataset with 300,000 video clips for action recognition. Similarly, we seek a suitable dataset for person search pre-training, but collecting a large-scale dataset is challenging especially when we need expensive person identification labels. Although unsupervised pre-training is a straightforward solution, it costs a large amount of computational resources, which are usually not affordable, for example,~\cite{up_reid} pre-train their UP-ReID models using 168 GPU days. Aiming to minimize computational costs for pre-training person search models, we attempt to use currently available datasets with or without labels for pedestrian detection and re-ID, which are sub-tasks for person search.

\begin{figure}[t]
	\centering

        \includegraphics[width=\linewidth]{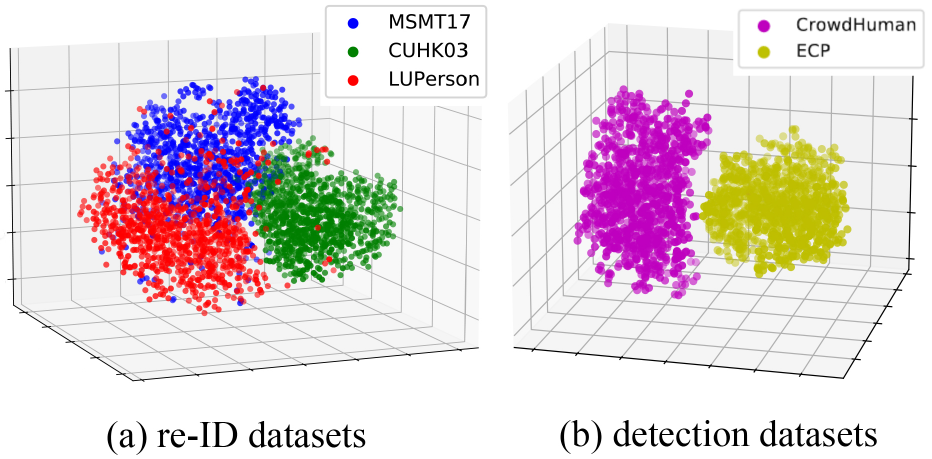}
	
	\caption{t-SNE visualization of features with ResNet50 pretrained on ImageNet from 1,000 random samples in different detection and re-ID datasets. 
 }
	\label{fig:tsne}
\end{figure}

  Since we use sub-task datasets for pre-training,
 two crucial considerations must be taken into account:
 (1) \textbf{How to support full-task training by sub-task data?} Pedestrian detection datasets only provide person bounding box (bbox) annotations that can only facilitate training for the detection sub-task; while re-ID datasets lack background information and thus can only facilitate training for the re-ID sub-task.  It is challenging to design a learning paradigm such that the entire person search model is well optimized.
(2) \textbf{How to deal with domain discrepancy under limited data?} 
Compared to other pre-training methods that use a large amount of data, the data we can use is rather limited. We observe that samples from different datasets, such as CrowdHuman~\cite{shao2018crowdhuman} and EuroCity Persons (ECP)~\cite{ecp}, exhibit significant differences in style appearance and underlying distributions (Fig.~\ref{fig:tsne}).
Thus, it is necessary to deal with domain discrepancy so as to enhance the generalization ability of our pre-trained models. 
 
To facilitate pre-training on sub-task data, we present a customized pre-training method. Specifically, we propose a hybrid training paradigm so that data with different kinds of supervisions can be handled in one joint framework; also, to alleviate the negative impact brought by domain discrepancy, we propose an intra-task alignment module (IAM), which is used to align features for detection and re-ID sub-tasks separately, so that the learned representations are domain invariant.
As shown in Fig.~\ref{fig:methods}, our pre-trained model significantly outperforms the ImageNet pre-trained model across different protocols (\emph{i.e.} methods, fine-tuning data, pre-training data and backbones) on the PRW dataset~\cite{PRW}. These results show that our approach provides stronger pre-trained models for the person search task.

In summary, our contributions can be summarized as follows:

  \begin{enumerate}
    \item Due to the lack of large-scale person search datasets, we propose to use off-the-shelf datasets of sub-tasks (pedestrian detection and re-ID task) for person search pre-training. To the best of our knowledge, this is the first work that investigates how to support full-task pre-training using sub-task data for person search. We believe our attempt is inspiring and encouraging for future work.
    \item We propose a novel pre-training method specific for person search. It consists of a hybrid learning paradigm that handles data with different kinds of supervisions, and an intra-task alignment module that alleviates domain discrepancy under limited resources.
    \item We provide analyses showing that our pre-training method is generalizable across different backbones and our pre-trained models benefit different methods.
    Our pre-trained models are more effective than the ImageNet ones and thus are expected to essentially contribute to the person search community.
    Benefited from our pre-training method, we establish a new state of the art on the CUHK-SYSU~\cite{CUHK} and PRW~\cite{PRW} benchmarks.
\end{enumerate}

\section{Related Work}

Since this paper develops pre-trained models specifically promoting person search, we thereby review relevant works from the above two aspects.

\subsection{Person Search}
Due to its wide application prospects, person search has made rapid progress in recent years. Some works~\cite{TCTS,OR,IGPN,FPN+RDLR,CNN+MGTS,CNN+CLSA} optimize the pedestrian detection and re-ID tasks in person search separately (two-stage methods), while others~\cite{CUHK,DAPS,CVPR_trans1,CGPS,CVPR_trans2,R-SiamNets,IJCV_NAE,PGA,SeqNet,TriNet,Alignps,Alignps_roi} regard two sub-tasks as a whole and jointly optimize them in an end-to-end manner (one-stage methods). Generally, existing person search methods can be roughly divided into two groups: fully supervised and weakly supervised methods.
\textbf{(1) \emph{Fully supervised.}} Given human bbox and their corresponding identity information, this group of methods allows for direct model training. As a pioneering method for person search,~\cite{CUHK} propose a joint framework that incorporates re-ID layers on top of Faster-RCNN~\cite{fasterrcnn} detector.~\cite{IJCV_NAE} introduce a norm-aware embedding method (NAE) to accommodate the conflicting optimization objectives of detection and re-ID.~\cite{Alignps} construct an anchor-free framework to address the misalignment in different levels (including scales, regions, and tasks). To improve performance, some recent methods~\cite{CVPR_trans1,CVPR_trans2} utilize transformer to learn more discriminative feature representations. \textbf{(2) \emph{Weakly supervised.}} Considering that obtaining identity information is more difficult, this group of methods only uses annotations of bbox for training.~\cite{R-SiamNets} propose a region siamese network for recognition in the absence of identity annotations.~\cite{CGPS} propose a weakly supervised person search method by leveraging context clues of detection, memory and scene in unconstrained natural images.

In this paper, we study person search from a novel perspective, making full use of unlabeled and labeled data of sub-tasks to develop powerful pre-trained models for person search.

\subsection{Vision Model pre-training}
Benefiting from the strong visual knowledge distributed in ImageNet~\cite{ImageNet}, fine-tuning a pre-trained model with a small amount of task-specific data can perform well on downstream tasks. This triggers the first wave of exploring pre-trained models in the era of deep learning.

The early efforts of pre-training are mainly achieved in a supervised manner. By applying ResNet~\cite{resnet} pre-trained on ImageNet as the backbone, various vision tasks (\emph{e.g.} image classification and segmentation) have been quickly advanced.
In comparison with supervised pre-training, self-supervised pre-training allows for huge advances. Some methods~\cite{luperson,luperson_noisy,Chenaug_ICLR} construct positive pairs with data augmentation, and obtain pre-trained models via contrastive learning~\cite{moco_v2}.~\cite{detreg} propose a self-supervised method that pre-trains the entire object detection network, including the object localization and embedding components.~\cite{mae} adopt an asymmetric encoder-decoder network and show that scalable vision learners can be obtained simply by reconstructing the missing pixels.~\cite{MaskFeat} use histograms of oriented gradients to learn abundant visual knowledge for pre-training video models.~\cite{luperson} study the key factors (\emph{i.e.} data augmentation and contrastive loss) to improve the generalization ability of learned re-ID features. Based on the contrastive learning pipeline,~\cite{up_reid} propose an unsupervised framework to learn the fine-grained re-ID features. Within a contrastive learning framework,~\cite{idfree} attempt to learn person similarity without using manually labeled identity annotations.

Compared to individual task pre-training, it is much more challenging to pre-train for a hybrid task like person search, due to the lack of large-scale datasets. In this paper, we investigate how to support full-task pre-training using sub-task data, which has not been studied by previous works and our findings are expected to be inspiring and encouraging for future works.

\section{Method}
In this section, we begin with an overview of our pre-training framework, followed by an explanation of the proposed hybrid learning approach. Due to  the domain discrepancy in the hybrid learning, we introduce our simple intra-task alignment module (IAM) in detail at last.

\subsection{Overview}

\begin{figure}[t]
	\centering
		\includegraphics[width=\linewidth]{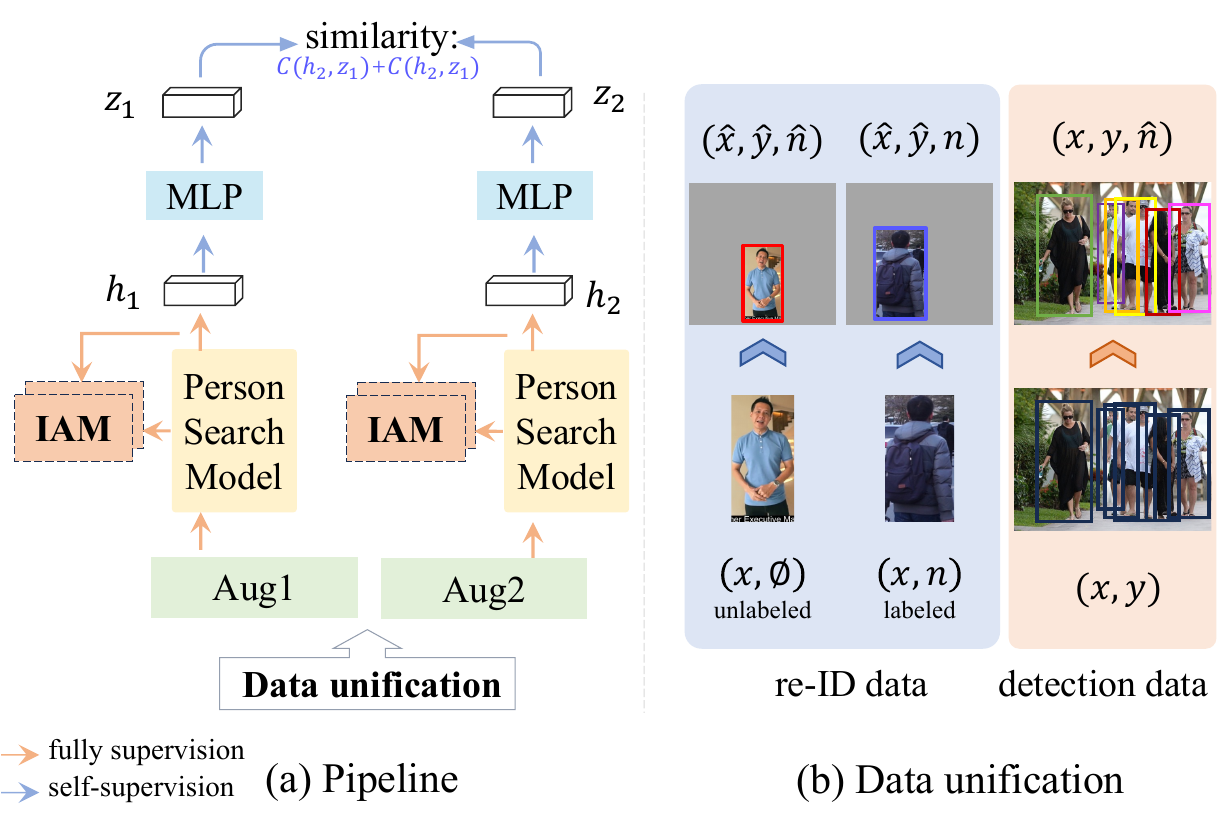}
	\caption{The pipeline of our pre-training method. MLP refers to a prediction multilayer perceptron. IAM is used to alleviate domain discrepancy across different datasets. 
 The right figure is the data unification operation to unify different data types for a unified interface. Black bboxes denote the people with no identity while colorful bboxes refer to persons with different identities.}
	\label{fig:views}
\end{figure}

In this paper, we propose a novel pre-training method, which distills specific knowledge from data of two sub-tasks to promote person search. In Fig.~\ref{fig:views}, we show the pipeline of our pre-training method, which is trained using a hybrid learning approach in an end-to-end manner. Following Simsiam~\cite{siamese}, our framework consists of a shared person search model, a prediction multilayer perceptron (MLP), a data unification module and IAMs. As shown in Fig.~\ref{fig:views}, we randomly sample images $x^D$ and $x^R$ from detection and re-ID datasets and perform data unification on them to handle different data types, where $x^D=\{x^{D_i}|D_i{\in}\{D_1, ..., D_M\}\}$ and $x^R=\{x^{R_j}|R_j{\in}\{R_1, ..., R_N\}\}$. Initially, we apply two data augmentations to each image, resulting in two different views of the same image.
Since we use sub-task datasets to train a person search model, specific flows are designed for different kinds of data:
(1) For $x$ from any dataset, each view goes through the entire person search model for detection task training and re-ID task training in a fully supervised way. (2) For $x$ from datasets without person identities, \emph{i.e.} unlabeled re-ID datasets and detection datasets, it is used for re-ID task training in a self-supervised way, \emph{i.e.} two views are sent to the framework for contrastive learning.
Moreover, two IAMs for detection and re-ID respectively, are developed to alleviate domain discrepancy by aligning features across different datasets.

\subsection{Hybrid Learning}
In a dilemma of lacking large-scale person search datasets, we use sub-task datasets for training. Each dataset only provides labels for each sub-task, or even no labels at all like the unlabeled re-ID datasets.
In order to handle different types of data in a unified framework and provide a simplified training pipeline, we propose a hybrid learning approach that combines self-supervised and fully-supervised learning to assist the unified framework in gathering knowledge from different datasets. The loss function for the entire pre-training procedure can be expressed as:
\begin{equation}
      L = L_{ps}+\eta L_{con}+\lambda L_{adv},
      \label{equ:total_loss}
\end{equation}
where $\lambda$ and $\eta$ are hyper-parameters; $L_{ps}$ (Eq.~\ref{equ:loss_ps}), $L_{con}$ (Eq.~\ref{equ:loss_con}) and $L_{adv}$ (Eq.~\ref{equ:loss_adv}) are introduced in the following.

\subsubsection{Data unification} 
The schematic diagram of data unification is shown in Fig.~\ref{fig:views} (b).

For pedestrian detection datasets, $D_i{=}(x^{D_i},y^{D_i})_{i=1}^{N_{D_i}}$ only provide position information of each person, without identity. $x$ and $y$ denote the image and person bbox annotations respectively; and ${N_{D_i}}$ is the total number of image samples in the corresponding dataset. To construct a unified interface and simplified pipeline, we regard each person in $D_i$ as different persons. Therefore, we have the detection dataset $D_i{=}(x^{D_i},y^{D_i},\hat{n}_i^{D_i})_{i=1}^{N_{D_i}}$, $\hat{n}$ is the new identity annotation.

For the re-ID datasets, the images form datasets are cropped from entire images, resulting in the absence of contextual background information. We obtain labeled re-ID dataset $R_j{=}(x_j^{R_j},n_j^{R_j})_{j=1}^{N_{R_j}}$ and the unlabeled re-ID dataset $R_j^{un}{=}(x_j^{R_j^{un}},)_{j=1}^{N_{R_j^{un}}}$ based on the presence of identity annotation. $x$ and $n$ denote the image and identity annotation. ${N_{R_j}}$, ${N_{R_j^{un}}}$ are the number of image samples in corresponding datasets. 
To construct a unified interface and simplified pipeline, similar to detection datasets, we also regard each person as a new individual. In addition, we randomly put the re-ID image on a canvas of varying proportions to the size of the image and resize to a fixed size. We refer to this operation as ``expand\_resize". Therefore, we have the labeled re-ID dataset $R_j{=}(\hat{x}_j^{R_j},\hat{y}_j^{R_j},n_j^{R_j})_{j=1}^{N_{R_j}}$ and unlabeled re-ID dataset $R_j^{un}{=}(\hat{x}_j^{R_j^{un}},\hat{y}_j^{R_j^{un}},\hat{n}_j^{R_j^{un}})_{j=1}^{N_{R_j^{un}}}$, where $\hat{x}$ denotes the canvas containing re-ID samples, $\hat{n}$ is new identity annotation, $\hat{y}$ is the location of each person on its canvas.

\subsubsection{Fully-supervised learning} 
After data unification, we can perform person search pre-training in a fully-supervised way. 
Given an arbitrary image $x$ and its person bbox annotations $y$, we optimize the encoder and detection head in person search model by the following detection loss function:
\begin{equation}
\begin{aligned}
      L_{det}= \sum\nolimits_{i=0}^J L_{rpn}(x_i,y_i)
      +L_{det-head}(x_i,y_i),
      \label{equ:loss_det}
\end{aligned}
\end{equation}
where $L_{rpn}$ indicates the loss for the RPN network~\cite{fasterrcnn}; $L_{det-head}$ is the loss for the detection head. $J$ is the image number of all sub-task datasets. 
With the image $x$ and its identity annotation $n$, we optimize the encoder and the re-ID head via the following re-ID loss function:
\begin{equation}
      L_{reid} = \sum\nolimits_{i=0}^J L_{oim}(x_i,n_i),
      \label{equ:loss_reid}
\end{equation}
where $L_{oim}$ is identical to that used by NAE~\cite{CUHK}. 
Please note person samples from $x^{D_i}$ and $x^{R^{un}}$ serve as different identities in $L_{oim}$~\cite{CUHK}, so that person representation learning can benefit from more unlabeled samples. 

Therefore, the total loss function for fully-supervised learning can be written as:
\begin{equation}
      L_{ps} = L_{det}+L_{reid}.
      \label{equ:loss_ps}
\end{equation}

\subsubsection{Self-supervised learning} 

In order to better distill re-ID knowledge from images without ID labels, we perform self-supervised learning for the re-ID sub-task.
Given an arbitrary image $x$ sampled from $\{D_1,..., D_M\}$ or $R^{un}$, we use two different augmentations to generate different views $Aug_1(x)$ and $Aug_2(x)$. Each view $Aug_i(x)$ goes through the entire person search model to obtain deep feature representations, denoted by $h_i= P(Aug_i(x))$, where $P$ is the entire person search model. And then $h_i$ is transformed to $z_i$ by the prediction MLP head. To maintain the consistency of predictions from two views, we minimize the negative cosine similarity as follows:

\begin{equation}
      C(h_2, z_1) = -\sum\nolimits_{k=0}^K \frac{h_2^k}{||h_2^k||_2} \cdot  \frac{z_1^k}{||z_1^k||_2},
      \label{equ:sim}
\end{equation}
where $||\cdot||_2$ is $l_2$ norm. $K$ refers to the image number of detection datasets and unlabeled re-ID dataset. Inspired by a recent work on contrastive learning~\cite{siamese}, we further adopt a symmetrized loss in self-supervised learning as,
\begin{equation}
      L_{con}(h, z) = \frac{1}{2}C(sg(h_2), z_1) \\
      + \frac{1}{2}C(sg(h_1), z_2),
      \label{equ:loss_con}
\end{equation}
where $sg$ denotes the operation of stop-gradient~\cite{siamese}. In this way, $P$ receives no gradient from $z_i$ in order to alleviate collapse solutions in siamese network during optimization.

\subsection{Intra-task Alignment Module}
\label{sec:IAM}
\begin{figure}[!t]
\begin{center}
   \includegraphics[width=1\linewidth]{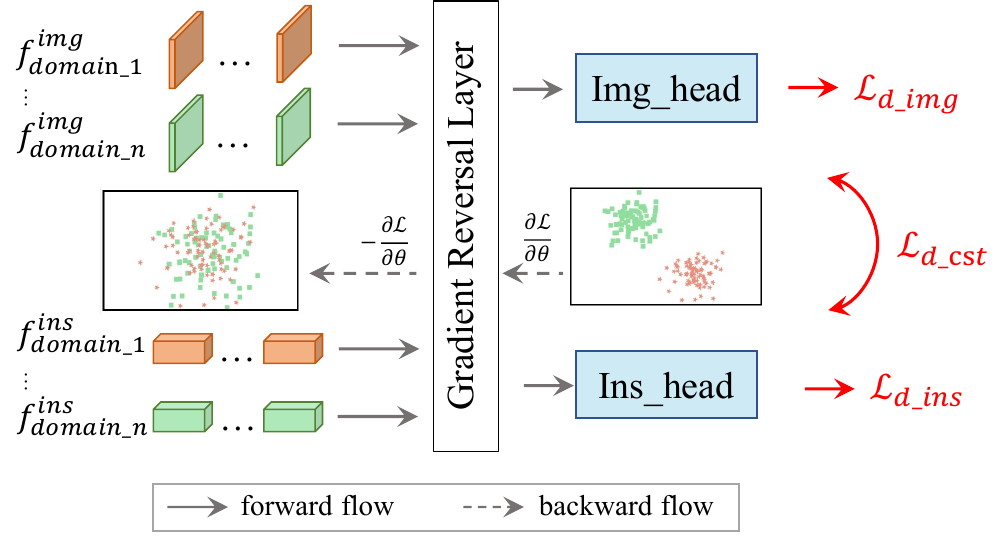}
\end{center}
   \caption{Structure of intra-task alignment module (IAM). Gradient reversal layer~\cite{grl} automatically reverses the gradient during backward propagation to achieve feature alignment across different datasets (domains) with the heads. In the backward flow, the colorful points refer to different features from different domains.}
\label{fig:intra}
\end{figure}

There are differences in data distribution among datasets of the same task (Fig. ~\ref{fig:tsne}), making it difficult to ensure effective distillation of task-related knowledge. To solve this problem, we present an IAM, which performs adversarial-based feature alignment at both instance and image levels. As shown in Fig.~\ref{fig:views}, we use two separate IAMs (\emph{i.e.} detection IAM and re-ID IAM) to reduce the discrepancy for pedestrian detection and re-ID datasets, respectively. The reason for such design is that pedestrian detection and re-ID conflict with each other in terms of features. Pedestrian detection task is concerned with identifying shared features among individuals, whereas re-ID aims to capture the distinct features and specific details of each identity.

The architecture of detection IAM and re-ID IAM is identical. As shown in Fig.~\ref{fig:intra}, we use features at both image level $f^{img}$ and instance level $f^{ins}$ as the input of IAM. For detection IAM, $f^{img}$ refers to the output feature maps of the encoder in the person search model, while $f^{ins}$ is the feature vectors from the detection head of the person search model. For re-ID IAM, $f^{img}$ is the feature maps extracted by the ROI-Align operation and $f^{ins}$ represents the feature vectors from the re-ID head in the person search model. Subsequently, these features are passed through several cascaded convolutional layers, followed by two fully connected (FC) layers termed as Img\_head and Ins\_head respectively. Finally, the predictions from two heads are used to perform domain classification based on $f^{img}$ and $f^{ins}$, which is optimized by the following loss function:

\begin{equation}
\begin{split}
L_{adv} &{=} L_{d\_{img}}(x^D,x^R,\overline{y})+L_{d\_{ins}}(x^D,x^R,\overline{y})\\
&+L_{d\_{cst}}(x^D,x^R),
\label{equ:loss_adv}
\end{split}
\end{equation}
where $\overline{y}$ indicates the domain label (\emph{i.e.} to distinguish
which dataset an image belongs to); $L_{d\_img}$ and $L_{d\_ins}$ are multi-class cross-entropy loss that applied to the image  level and instance level respectively. We note that maintaining the consistency between the domain classifiers at both levels is helpful to improve performance~\cite{dafasterrcnn} (from which we
observe a 0.3pp increase w.r.t. mAP.). To this end, we introduce a consistency regularizer $L_{d\_{cst}}$, following~\cite{dafasterrcnn}. 
After applying our IAM, we obtain domain-invariant features that are agnostic to the specific data domain.

\section{Experiments}

In this section, we first introduce benchmark datasets and evaluation metrics. Then, a series of experimental analyses are conducted. Finally, we compare with state-of-the-art methods on multiple benchmarks. 
Implementation details are provided in our code.
\subsection{Datasets and Evaluation Metrics}
\textbf{Datasets for pre-training.} We use two relatively large person detection datasets, \emph{i.e.} CrowdHuman~\cite{shao2018crowdhuman} and EuroCity Persons (ECP)~\cite{ecp} datasets. In addition, we use two relatively common re-ID datasets including MSMT17~\cite{msmt17} and CUHK03~\cite{cuhk03}, and one unlabeled large re-ID dataset, LUPerson~\cite{luperson} for pre-training. 

\textbf{Person search datasets.} The two most commonly used datasets for person search are PRW~\cite{PRW} and CUHK-SYSU~\cite{CUHK}. \textbf{CUHK-SYSU} is a large-scale person search dataset, providing 18,184 images and 8,432 individuals, with some images sourced from movie snapshots and others from street/city scenes. The training set consists of 11,206 images and 5,532 identities, while the testing data contains 6,978 gallery images with 2,900 query persons. Instead of using the entire testing set as a gallery, we follow the standard protocols with gallery sizes ranging from 50 to 4,000. We use the default gallery size of 100 in our experiments unless otherwise specified. \textbf{PRW} includes 11,816 images captured with six cameras on a university campus. It provides 5,134 training images with 482 identities while the testing set includes 2,057 different query persons and 6,112 gallery images. We use all gallery images as the search space for each query person. In addition to these two standard datasets, there is a new dataset, \textbf{PoseTrack21}~\cite{psetrack21}, that can be used for person search. It consists of 42,861 training images with 5,474 different individuals, and 19,935 gallery images with 1,313 query individuals. Unlike the above datasets, the queries in PoseTrack21 may contain multiple individuals in cases of occlusion.

\textbf{Evaluation Metrics.} mAP and Top-$k$ cumulative matching characteristics are the performance metrics for person search. The mAP metric refers to the accuracy and matching rate of searching a query person from the gallery images. The Top-$k$ score reflects the percentage of queries for which at least one of the $k$ most similar proposals succeeds in the re-ID matching step.

\subsection{Analysis}
In this section, we conduct a series of analyses on our method. Unless otherwise specified, our pre-training method is trained on five datasets with a total of 106,784 images, including CrowdHuman, ECP, MSMT17, CUHK03 and LUPerson30k datasets. All experiments in this subsection use PRW as the target dataset for verification unless otherwise specified.

\subsubsection{Generalizable to different backbones}

\begin{table}[t]
\begin{center}
\begin{tabular}{l|cc}
\hline\noalign{\smallskip}
Backbone   & mAP & Top-1\\
\noalign{\smallskip}
\hline
\noalign{\smallskip}
ResNet50            &42.67$\rightarrow$47.08 ($\uparrow$10.3\%)	  & 81.33$\rightarrow$84.00\\
ResNet18          &35.87$\rightarrow$39.79 ($\uparrow$10.9\%)   &77.43$\rightarrow$79.76 \\
ShuffleNet    &34.29$\rightarrow$40.98 ($\uparrow$19.5\%)  &78.07$\rightarrow$80.74 \\
\noalign{\smallskip}
\hline
\end{tabular}
\caption{Performance gain of NAE~\cite{NAE} with different pre-trained backbones.
A$\rightarrow$B: A refers to the performance using ImageNet pre-trained model, while B is that using our pre-trained model. The number after $\uparrow$ is relative improvement.}
\label{table:backbone}
\end{center}
\end{table}

We apply our pre-training method on top of  different backbones: ResNet18~\cite{resnet}, ResNet50~\cite{resnet}, ShuffleNet~\cite{zhang2018shufflenet}, and provide three pre-trained models to a method NAE.
As shown in Tab.~\ref{table:backbone}, compared to ImageNet pre-trained models, our pre-trained models achieve more than 10$\%$ relative improvements w.r.t. mAP across three different backbones, and especially the gain is up to 19.5$\%$ for ShuffleNet. These consistently significant improvements demonstrate our pre-training approach well generalizes to various backbones.

\subsubsection{Beneficial for different methods and datasets}
We also feed our pre-trained model (ResNet50) to different person search methods for initialization, including an anchor-based method (NAE~\cite{NAE}), an anchor-free method (AlignPS~\cite{Alignps}) and ROI-AlignPS~\cite{Alignps_roi} (mixing NAE and AlignPS).
As shown in Tab.~\ref{table:settings}, 
all methods improve significantly over its counterpart using ImageNet pre-trained model, w.r.t. both mAP and Top-1 accuracy.
It is notable that NAE surpasses AlignPS when switched to our pre-trained model.
Even for a very strong method ROI-AlignPS, the improvement is still impressive, \emph{i.e.} $\sim$3pp w.r.t. mAP.
These results indicate that our pre-trained models are beneficial to various methods.

Due to ROI-AlignPS~\cite{Alignps_roi} is the current available state-of-the-art one-stage method, we use it as baseline and fine-tune it on different datasets. As shown in Tab.~\ref{table:diff_ds}, the results with our pre-trained models are better than those with ImageNet pre-trained models.

\begin{figure}[!t]
  \centering
  \includegraphics[width=\linewidth]{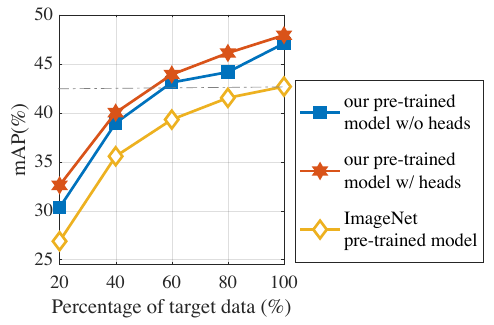}
  \caption{Performance comparison for NAE~\cite{NAE} with different pre-trained models on PRW dataset. By using only 60\% of target data, our pre-trained model outperforms the ImageNet pre-trained model trained on all annotations. 
  }
  \label{fig:scale}
\end{figure}

\setlength{\tabcolsep}{2.5pt}
\begin{table}[!t]
\begin{center}
\begin{tabular}{l|cc}
\hline\noalign{\smallskip}
Method  & mAP & Top-1\\
\noalign{\smallskip}
\hline
\noalign{\smallskip}

   NAE      &    42.67$\rightarrow$47.08 ($\uparrow$4.41)	  & 81.33$\rightarrow$84.00 ($\uparrow$2.67)\\
  AlignPS     & 45.90$\rightarrow$49.14 ($\uparrow$3.24)  & 81.90$\rightarrow$83.56 ($\uparrow$1.66)\\
    ROI-AlignPS & 51.84$\rightarrow$54.47 ($\uparrow$2.63)  & 85.48$\rightarrow$86.56  ($\uparrow$1.08)\\
\noalign{\smallskip}
\hline
\end{tabular}
\caption{Performance gain of different person search methods using our pre-trained model instead of ImageNet model. 
A$\rightarrow$B: A refers to the performance using ImageNet pre-trained model, while B is that using our pre-trained model.
}
\label{table:settings}
\end{center}
\end{table}
\setlength{\tabcolsep}{1.4pt}

\setlength{\tabcolsep}{2.5pt}
\begin{table}[!t]
\begin{center}
\begin{tabular}{l|cc}
\hline\noalign{\smallskip}
Dataset  & mAP & Top-1\\
\noalign{\smallskip}
\hline
\noalign{\smallskip}
  PRW     &    51.84$\rightarrow$54.47	  & 85.48$\rightarrow$86.56\\

  CUHK-SYSU     & 95.33$\rightarrow$95.38  & 95.76$\rightarrow$96.03\\
  
  PoseTrack21 & 59.21$\rightarrow$63.62  & 82.10$\rightarrow$87.13\\
\noalign{\smallskip}
\hline
\end{tabular}
\caption{Performance gain of ROI-AlignPS~\cite{Alignps_roi} on different datasets using our pre-trained model instead of ImageNet model. 
A$\rightarrow$B: A refers to the performance using ImageNet pre-trained model, while B is that using our pre-trained model.
}
\label{table:diff_ds}
\end{center}
\end{table}
\setlength{\tabcolsep}{1.4pt}

\subsubsection{Requiring less target training data}
We observe the performance of NAE when less target training data is used for fine-tuning.
From Fig.~\ref{fig:scale}, we have the following observations: 
(1) When using our pre-trained model, it outperforms its counterpart using the ImageNet pre-trained model by a large margin when different amounts of target data are used for fine-tuning. 
(2) Our pre-trained NAE surpasses its ImageNet pre-trained counterpart, when using only 60\% of the target data.
(3) When we initialize the entire model (including both backbones and heads) with our pre-trained model, the performance gain is larger than that solely initializes backbone, parameters, which is consistent with~\cite{detreg}.
These results demonstrate our pre-training model provides a better initialization so that the person search method requires much less target data for fine-tuning to achieve comparable performance.

\subsubsection{Effect of re-ID data operation in data unification}
Multi-scale matching problem is an underlying challenge of person search~\cite{multi-scale2018eccv}. We obtain a scale-invariant feature through pre-training by placing re-ID images on canvases of different proportions and resizing them to a fixed size, which helps alleviate the matching problem. We refer to our approach as ``expand\_resize" and the operation of randomly pasting re-ID images onto a fixed-size canvas as ``random\_paste". As shown in Tab.~\ref{table:aug}, we validate the effectiveness of ``expand\_resize".

\setlength{\tabcolsep}{4pt}
\begin{table}[!t]
\begin{center}
\begin{tabular}{l|cc}
\hline\noalign{\smallskip}
Operation    & mAP & Top-1\\
\noalign{\smallskip}
\hline
\noalign{\smallskip}
random\_paste       &46.49	    & 83.36\\

expand\_resize      &\textbf{47.08}	  & \textbf{84.00}\\
\noalign{\smallskip}
\hline
\end{tabular}
\caption{NAE performance~\cite{NAE} of using different re-ID data operation in data unification. ``random\_paste": paste a re-ID image randomly on a fixed size canvas. ``expand\_resize": put re-ID image on a canvas of different ratios of original image and resize to a fixed size.}
\label{table:aug}
\end{center}
\end{table}
\setlength{\tabcolsep}{1.4pt}

\subsubsection{Impact of our IAM}

\begin{figure}[!t]
\setlength{\abovecaptionskip}{1mm} 
  \centering
  \includegraphics[width=\linewidth]{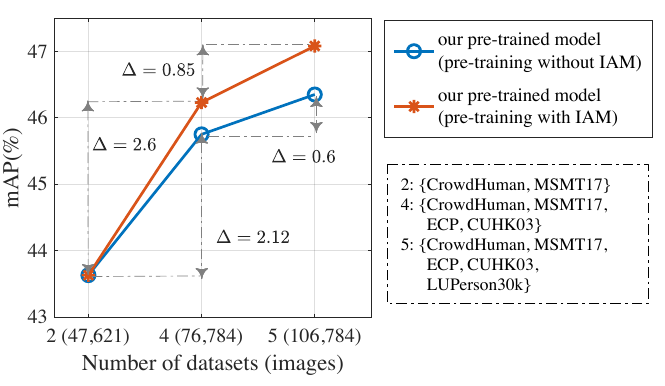}
  \caption{Effect of IAM with increasing pre-training datasets.
  }
  \label{fig:ds_number}
\end{figure}

We further investigate the effect of our proposed IAM module, which alleviates domain discrepancy by feature alignment.
From Fig.~\ref{fig:ds_number}, we can see that: (1) When we use more datasets for pre-training, the performance gain becomes smaller, from 2.12pp for 4 datasets to 0.6pp for 5 datasets.
(2) By employing our IAM, the performance gain grows at different number of datasets.

It is possible to perform both intra-task and inter-task alignment to alleviate domain discrepancy. We study their effects by adding them one by one. From Tab.~\ref{table:components}, we can see that intra-task alignment achieves 47.08pp w.r.t. mAP, yet inter-task alignment shows a negative effect by dropping the performance slightly. The reason is that different tasks are of different optimization goals, and thus it is unfavorable to align the features across two tasks. Based on the above observations, we only use intra-task alignment in our method.

\begin{table}[!t]
\begin{center}
\begin{tabular}{p{3.4cm}<{\centering}p{2.5cm}<{\centering}|cc}
\hline\noalign{\smallskip}
intra-task AMs (IAMs) & inter-task AM  & mAP & Top-1\\
\noalign{\smallskip}
\hline
\noalign{\smallskip}
\checkmark &              &\textbf{47.08}	  & \textbf{84.00}\\

\checkmark& \checkmark    &46.53	  & 83.16\\
\noalign{\smallskip}
\hline
\end{tabular}
\caption{Effects of intra-task alignment module (IAM) and inter-task alignment module.}
\label{table:components}
\end{center}
\end{table}
\begin{table}[!t]
\begin{center}
\begin{tabular}{c|l|cc|cc}
\hline\noalign{\smallskip}
\multicolumn{2}{c}{\multirow{2}{*}{Methods}}
 & \multicolumn{2}{|c}{CUHK-SYSU} & \multicolumn{2}{|c}{PRW}\\
 \noalign{\smallskip}
\cline{3-6} 
\noalign{\smallskip}
\multicolumn{2}{c|}{} &mAP & Top-1 & mAP & Top-1   \\   
\noalign{\smallskip}
\hline
\noalign{\smallskip}
\multirow{7}{*}{\rotatebox{90}{two-stage}}
&DPM+IDE               & -     & -    & 20.5 & 48.3\\
&CNN+CLSA         & 87.2  & 88.5 & 38.7 & 65.0\\
&FPN+RDLR         & \underline{93.0}  & \underline{94.2} & 42.9 & 70.2\\
&IGPN                 & 90.3  & 91.4 & \underline{47.2} & \underline{87.0}\\
&OR                    & 92.3  & 93.8 &\textbf{52.3} & 71.5\\
&TCTS                 & \textbf{93.9}  & \textbf{95.1} & 46.8 & \textbf{87.5}\\
\noalign{\smallskip}
\hline 
\hline 
\noalign{\smallskip}
\multirow{20}{*}{\rotatebox{90}{one-stage}}
&OIM                  & 75.5  & 78.7 & 21.3 & 49.4\\

&NPSM                 & 77.9  & 81.2 & 24.2 & 53.1\\
&RCAA                  & 79.3  & 81.3 & -    & -   \\
&CTXG                  & 84.1  & 86.5 & 33.4 & 73.6\\
&QEEPS                & 88.9  & 89.1 & 37.1 & 76.7\\
&HOIM                  & 89.7  & 90.8 & 39.8 & 80.4\\
&BINet                & 90.0  & 90.7 & 45.3 & 81.7\\
&NAE\dag                & 91.5  & 92.4 & 42.7 & 81.3\\
&PGA                    & 92.3  & 94.7 & 44.2 & 85.2\\
&SeqNet              & 93.8  & 94.6 & 46.7 & 83.4\\

&AGWF              & 93.3  & 94.2 & \underline{53.3} &\underline{87.7} \\
&AlignPS           & 93.1  & 93.4 & 45.9 & 81.9\\
&PSTR                & 93.5  & 95.0 & 49.5 & \textbf{87.8}\\
&COAT             & 94.2  & 94.7 & 45.9 & 81.9\\

&ROI-AlignPS\dag   & \underline{95.3}  & \underline{95.8} & 51.8 & 85.5\\
\noalign{\smallskip}
\cline{2-6} 
\noalign{\smallskip}

&ROI-AlignPS w/ ours   & \textbf{95.4} & \textbf{96.0} & \textbf{54.5} & 87.6\\   
\noalign{\smallskip}
\hline
\end{tabular}

\caption{Comparison with state-of-the-art methods on CUHK-SYSU and PRW datasets. Best results are bold and the second results are underlined. \dag~refers to our re-implementation.}
\label{table:sota}
\end{center}
\end{table}

\subsection{Comparison with SOTA Methods}
In this section, the backbones of all the methods are the ResNet50. We compare our method with the one-stage methods OIM~\cite{CUHK}, NPSM~\cite{NPSM}, RCAA~\cite{RCAA}, CTXG~\cite{CTXG}, QEEPS~\cite{QEEPS}, HOIM~\cite{HOIM}, BINet~\cite{BINet}, NAE~\cite{NAE}, PGA~\cite{PGA}, SeqNet~\cite{SeqNet}, AGWF~\cite{TriNet}, AlignPS~\cite{Alignps}, PSTR~\cite{CVPR_trans1}, COAT~\cite{CVPR_trans2}, ROI-AlignPS~\cite{Alignps_roi} and two-stage methods DPM+IDE~\cite{PRW}, CNN+CLSA~\cite{CNN+CLSA}, FPN+RDLR~\cite{FPN+RDLR}, IGPN~\cite{IGPN}, OR~\cite{OR}, TCTS~\cite{TCTS} on the common CUHK-SYSU dataset~\cite{CUHK} and PRW dataset~\cite{PRW}.

\textbf{Results on CUHK-SYSU and PRW datasets}. As shown in Tab.~\ref{table:sota}, we achieve 95.4pp and 96.0pp w.r.t. mAP and Top-1 scores respectively, establishing a state-of-the-art performance on CUHK-SYSU dataset. Due to the large size of the CUHK-SYSU dataset and the relatively small number of gallery samples for each query, it is relatively easy to achieve saturation performance on this dataset. As a result, our method only surpasses the baseline (\emph{i.e}. ROI-AlignPS\dag) by 0.1pp. In contrast, the PRW dataset has less training data and a larger gallery, resulting in a significant performance degradation. However, our method achieves 54.5pp w.r.t. mAP on the PRW dataset, obtaining an improvement of 2.7pp over the baseline ROI-AlignPS\dag.

\section{Conclusion}
In this work, we focus on designing a specific pre-training method for the person search task.
Considering the lack of large-scale person search datasets, we employ its sub-task (pedestrian detection and re-ID) datasets for pre-training and propose a unified framework to handle those datasets with and without labels, as well as large domain gaps.
Specifically, our proposed method consists of a hybrid learning paradigm that handles data with different kinds of supervisions, and an intra-task alignment module that alleviates domain discrepancy under limited resources.
We validate the effectiveness of our approach by providing insightful analyses from different perspectives. Additionally, we provide better pre-trained models than ImageNet ones for the person search community, which can be simply loaded by various methods for initialization to achieve higher performance. We use 106,784 images (20$\times$ larger than the data in the PRW dataset) for pre-training. As shown in Fig.~\ref{fig:ds_number}, the performance of our approach continues to improve with an increasing number of pre-training images. We will use more data to pre-train a more powerful model in the future.

\section{Acknowledgments}
This work was supported in part by the National Natural Science Foundation of China (Grant No. 62322602, 62172225, 62372077), Natural Science Foundation of Jiangsu Province, China (Grant No. BK20230033), CAAI-Huawei MindSpore Open Fund and the China Postdoctoral Science Foundation (Grant No. 2022M720624).

\bibliography{aaai24}

\end{document}